\pdfoutput=1

\documentclass{article}

\usepackage[preprint]{wenxiaobai}

\usepackage[utf8]{inputenc}
\usepackage[T1]{fontenc}
\usepackage{hyperref}
\usepackage{url}
\usepackage{booktabs}
\usepackage{amsfonts}
\usepackage{amsmath}
\usepackage{amssymb}
\usepackage{microtype}
\usepackage{xcolor}

\usepackage{graphicx}
\usepackage{multirow}
\usepackage{colortbl}
\usepackage{pifont}
\usepackage{algorithm}
\usepackage{algpseudocode}
\usepackage{tabularx}
\usepackage{wrapfig}

\graphicspath{{figures/}}

\definecolor{aragcolor}{HTML}{F5F4EB}
\definecolor{promptcolor}{HTML}{F8F4EF}
\definecolor{toolcolor}{HTML}{EEF2F8}
\definecolor{vanillacolor}{HTML}{EBEBEB}
\definecolor{graphworkflowcolor}{HTML}{CEDCE9}

\newtcolorbox{promptbox}[1][]{
  colback=white,
  colframe=gray!75!black,
  colbacktitle=promptcolor,
  coltitle=black,
  title=#1,
  fonttitle=\bfseries\small,
  fontupper=\small\ttfamily,
  boxrule=0.8pt,
  arc=2pt,
  left=4pt, right=4pt, top=4pt, bottom=4pt
}

\usepackage{tcolorbox}
\tcbuselibrary{listings,breakable}
\newtcolorbox{toolbox}[1][]{
  colback=white,
  colframe=blue!60!black,
  colbacktitle=toolcolor,
  coltitle=black,
  title=#1,
  fonttitle=\bfseries\small,
  fontupper=\small\ttfamily,
  boxrule=0.8pt,
  arc=2pt,
  left=4pt, right=4pt, top=4pt, bottom=4pt
}

\title{A-RAG: Scaling Agentic Retrieval-Augmented Generation via Hierarchical Retrieval Interfaces}

\author{
  \bf Mingxuan Du\textsuperscript{\rm 1}, Benfeng Xu\textsuperscript{\rm 2\textdagger}, Chiwei Zhu\textsuperscript{\rm 1}, Shaohan Wang\textsuperscript{\rm 1}, Pengyu Wang\textsuperscript{\rm 1} \\
  \bf Xiaorui Wang\textsuperscript{\rm 2}, Zhendong Mao\textsuperscript{\rm 1\textdaggerdbl} \\
  \textsuperscript{1}University of Science and Technology of China, Hefei, China \\
  \textsuperscript{2}Metastone Technology, Beijing, China \\
  \texttt{dumingxuan@mail.ustc.edu.cn}
}

\begin{document}
\maketitle

\begingroup
\renewcommand{\thefootnote}{\fnsymbol{footnote}}
\footnotetext[2]{Project lead.}
\footnotetext[3]{Corresponding author.}
\endgroup
\setcounter{footnote}{0}

\begin{abstract}
\hspace{2em}Frontier language models have demonstrated strong reasoning and long-horizon tool-use capabilities. However, existing RAG systems fail to leverage these capabilities. They still rely on two paradigms: (1) designing an algorithm that retrieves passages in a single shot and concatenates them into the model's input, or (2) predefining a workflow and prompting the model to execute it step-by-step. Neither paradigm allows the model to participate in retrieval decisions, preventing efficient scaling with model improvements. In this paper, we introduce \textbf{A-RAG}, an \underline{\textbf{A}}gentic \underline{\textbf{RAG}} framework that exposes hierarchical retrieval interfaces directly to the model. A-RAG provides three retrieval tools: keyword\_search, semantic\_search, and chunk\_read, enabling the agent to adaptively search and retrieve information across multiple granularities. Experiments on multiple open-domain QA benchmarks show that A-RAG consistently outperforms existing approaches with comparable or lower retrieved tokens, demonstrating that A-RAG effectively leverages model capabilities and dynamically adapts to different RAG tasks. We further systematically study how A-RAG scales with model size and test-time compute. We will release our code and evaluation suite to facilitate future research. Code and evaluation suite are available at \url{https://github.com/Ayanami0730/arag}.
\end{abstract}

\begin{wrapfigure}{r}{0.4\linewidth}
  \centering
  \vspace{-15pt}
  \includegraphics[width=\linewidth]{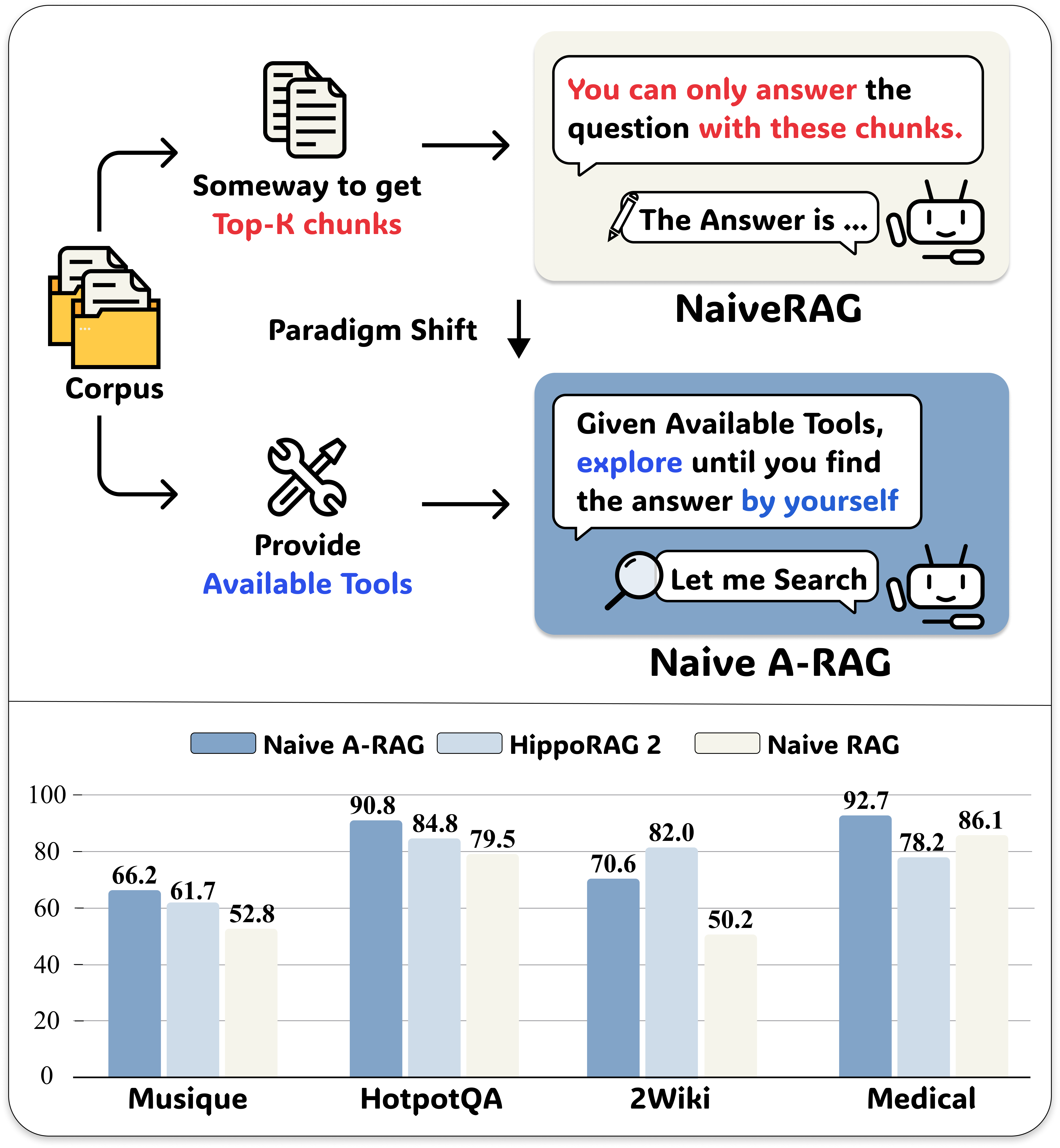}
  \caption{Two paradigms comparison and performance results.}
  \label{fig:paradigm-overview}
  \vspace{-15pt}
\end{wrapfigure}

\section{Introduction}

The development of LLMs has entered a new phase, where the primary scaling direction is shifting from single-turn text understanding and generation toward complex reasoning and multi-step, tool-augmented interaction~\citep{openai2025gpt5, anthropic2025claude45sonnet, google2025gemini3, deepseekai2025deepseekr1incentivizingreasoningcapability,shao2025deepseekmathv2selfverifiablemathematicalreasoning,yang2025qwen3technicalreport,kimiteam2025kimik2openagentic,minimax2025m2}. This transformation has significantly enhanced the capabilities and practicality of LLM-based agents, demonstrating remarkable progress in domains such as coding and deep research. By integrating frontier models, coding agents~\citep{cursor_ide, claude_code} have substantially improved the productivity of software engineers, while deep research agents~\citep{chatgpt_deep_research, gemini_deep_research, tongyideepresearchteam2025tongyideepresearchtechnicalreport} have greatly accelerated researchers' ability to conduct surveys and gather information. This marks a paradigm shift. However, methods in the RAG domain have rarely addressed this transition.

Existing RAG methods primarily rely on two paradigms: (1) designing an algorithm (with or without graph structures) that retrieves multiple passages in a single shot and concatenates them into the model's input~\citep{yan2024correctiveretrievalaugmentedgeneration,sarthi2024raptorrecursiveabstractiveprocessing,gutiérrez2025hipporagneurobiologicallyinspiredlongterm,edge2025localglobalgraphrag,guo2025lightragsimplefastretrievalaugmented,qian2025memoragboostinglongcontext, huang2025retrievalaugmentedgenerationhierarchicalknowledge}; (2) predefining a workflow and prompting the model to execute it step-by-step through multiple iterations~\citep{jiang2023activeretrievalaugmentedgeneration,trivedi2023interleavingretrievalchainofthoughtreasoning,asai2023selfraglearningretrievegenerate,liu2024raisflearninganswerunderstand}. Neither approach is truly agentic, as the model is not allowed to adapt the workflow based on the specific task, choose different interaction strategies, or decide when sufficient evidence has been gathered to provide an answer.

As illustrated in Figure~\ref{fig:paradigm-overview}, the key distinction between Naive RAG and Naive Agentic RAG lies in the agent's autonomy, and our preliminary experiments show that even the simplest Naive Agentic RAG, equipped with only a single embedding-based tool to retrieve from the corpus, consistently outperforms Naive RAG and previous baselines. This result demonstrates the potential of the agentic RAG paradigm.

To address these limitations, we propose A-RAG, an Agentic RAG framework featuring hierarchical retrieval interfaces. Our key insight is that information within a corpus is inherently organized at multiple granularities, ranging from fine-grained keyword-level signals to coarser sentence-level and chunk-level representations. Accordingly, we design a suite of retrieval tools that enable the agent to access information across these granularities. We observe that when equipped with this hierarchical toolset, the agent spontaneously generalizes to diverse workflows tailored to various tasks, yielding consistent performance gains.

Comprehensive experiments across multiple benchmarks demonstrate that A-RAG substantially surpasses prior methods. Furthermore, we conduct systematic studies on Test-Time Scaling behavior, showing that A-RAG's performance improves steadily with increased computational resources, indicating that our framework scales efficiently alongside advances in model capabilities. In summary, our contributions include:
\begin{itemize}
    \item We identify the paradigm shift from static LLM pipelines to dynamic agent-based systems, and highlight the necessity of transforming RAG into an agentic framework.
    \item We introduce A-RAG, an agentic RAG framework with hierarchical retrieval interfaces. By conducting comprehensive experiments, we validate that multi-granularity tools are essential for unlocking stronger model performance.
    \item We present further scaling analyses across multiple dimensions, demonstrating that our framework scales efficiently alongside advances in model capabilities and test-time computation.
\end{itemize}

\section{Related Work}

\begin{figure}[!htbp]
  \centering
  \includegraphics[width=0.95\linewidth]{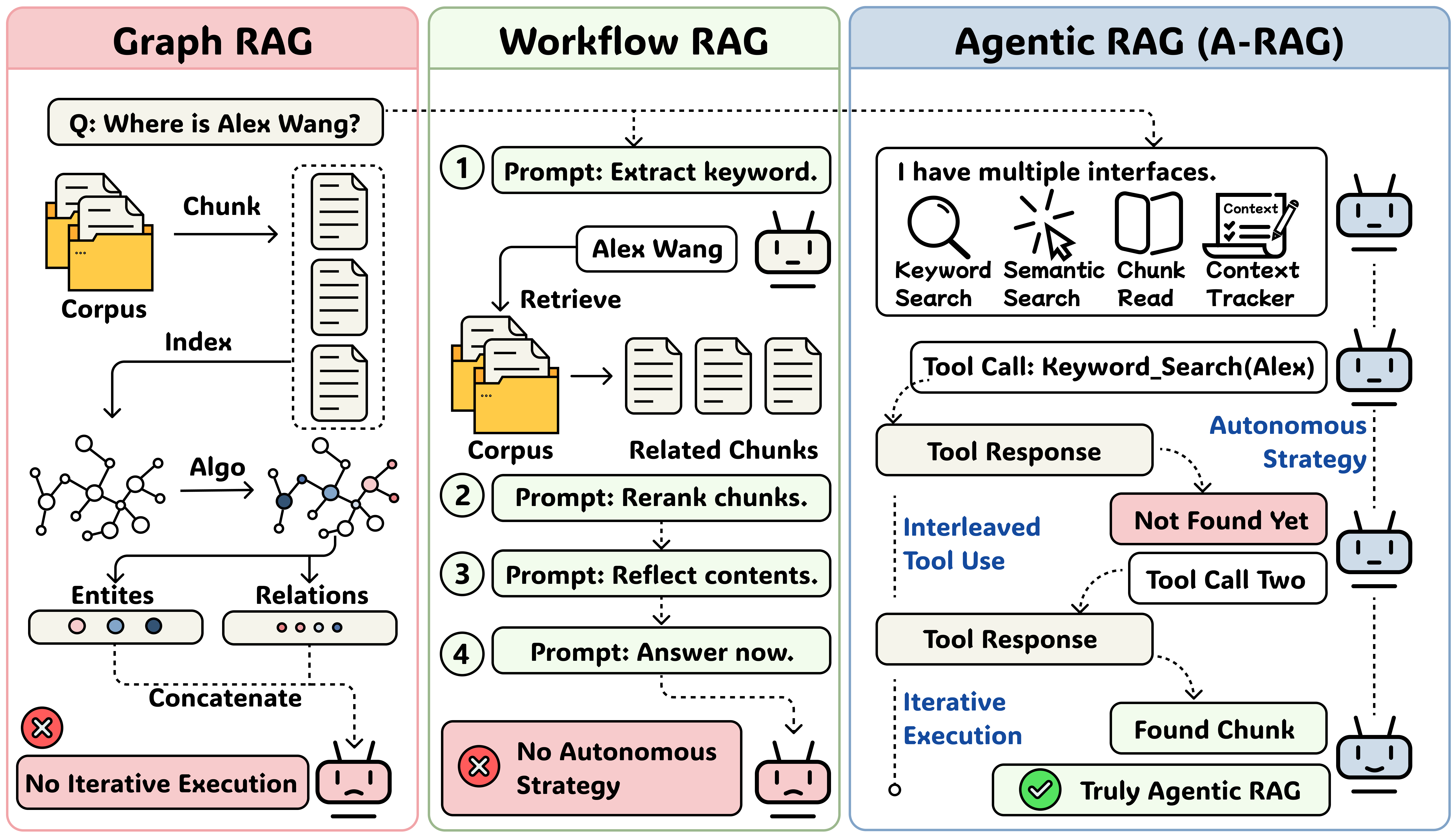}
  \caption{Comparison of three paradigms. We identify three principles of agentic autonomy: Autonomous Strategy, Iterative Execution, and Interleaved Tool Use. Only A-RAG satisfies all three, making it a truly agentic framework.}
  \label{fig:three-paradigms}
\end{figure}

We compare three RAG paradigms in Figure~\ref{fig:three-paradigms}: Graph RAG, Workflow RAG, and Agentic RAG (A-RAG). We identify three principles that define true agentic autonomy, and demonstrate that A-RAG is the only paradigm satisfying all three. A detailed comparison across existing methods is provided in Appendix~\ref{sec:autonomy-comparison}.

\subsection{Basic RAG}
Early research demonstrated that retrieval can help models incorporate external knowledge to answer questions more accurately~\citep{lewis2021retrievalaugmentedgenerationknowledgeintensivenlp}. Subsequent work has continuously improved upon this foundation through query rewriting~\citep{chan2024rqraglearningrefinequeries}, adaptive routing strategies~\citep{jeong2024adaptiveraglearningadaptretrievalaugmented}, retrieval quality evaluation~\citep{yan2024correctiveretrievalaugmentedgeneration}, and reranking mechanisms.

\subsection{Graph RAG}
In 2024, Microsoft introduced GraphRAG~\citep{edge2025localglobalgraphrag}, which constructs entity-relation graphs from corpora to help models develop holistic understanding of large-scale knowledge bases. This approach has rapidly evolved into a mainstream RAG paradigm, with researchers advancing the frontier through innovations in knowledge graph structure design, semantic unit definition, and retrieval strategies~\citep{guo2025lightragsimplefastretrievalaugmented,shen2025geargraphenhancedagentretrievalaugmented,yang2025graphsearchagenticdeepsearching,song2025efficienttransferableagenticknowledge}. Among these, RAPTOR~\citep{sarthi2024raptorrecursiveabstractiveprocessing} constructs hierarchical tree structures through recursive summarization for multi-level retrieval. LightRAG~\citep{guo2025lightragsimplefastretrievalaugmented} combines knowledge graphs with vector retrieval for both local and global search. HippoRAG~\citep{gutiérrez2025hipporagneurobiologicallyinspiredlongterm, gutiérrez2025ragmemorynonparametriccontinual} mimics hippocampal memory indexing using Personalized PageRank for efficient multi-hop reasoning. While these methods incorporate richer structure, they still rely on predefined retrieval algorithms rather than model-driven decisions. If the initially retrieved context is insufficient, the model cannot leverage its reasoning capabilities to iteratively gather more comprehensive and accurate information.

\subsection{Workflow RAG}
With the emergence of LLM-based agents, many works have explored agentic approaches to RAG. However, most rely on predefined agent-workflows that prompt models to execute fixed procedures step by step. So we refer to these methods as Workflow RAG. Some further employ SFT and RL to help models follow these workflows more robustly. Among the training-free methods, FLARE~\citep{jiang2023activeretrievalaugmentedgeneration} triggers retrieval when generation confidence drops, IRCoT~\citep{trivedi2023interleavingretrievalchainofthoughtreasoning} interleaves chain-of-thought reasoning with retrieval steps, and RA-ISF~\citep{liu2024raisflearninganswerunderstand} decomposes complex queries through iterative self-feedback. Multi-agent approaches further extend this paradigm: MA-RAG~\citep{nguyen2025maragmultiagentretrievalaugmentedgeneration} coordinates specialized agents via collaborative chain-of-thought, RAGentA~\citep{besrour2025ragentamultiagentretrievalaugmentedgeneration,chang2024mainragmultiagentfilteringretrievalaugmented} combines hybrid retrieval with citation tracking for question answering. Training-based methods have demonstrated that even smaller models can learn effective retrieval strategies~\citep{asai2023selfraglearningretrievegenerate,chan2024rqraglearningrefinequeries,chen2025improvingretrievalaugmentedgenerationmultiagent,xiong2025raggymsystematicoptimizationlanguage,jin2025searchr1trainingllmsreason,song2025r1searcherincentivizingsearchcapability,luo2025graphr1agenticgraphragframework}. Despite their sophistication, these workflows remain fixed at design time: the model cannot adapt its strategy based on task characteristics. In contrast, we demonstrate that with agent-friendly hierarchical retrieval interfaces, models can autonomously adopt diverse interaction strategies without any predefined workflows, exhibiting stronger and more robust performance across varying task complexities.

\begin{figure}[!htbp]
  \centering
  \includegraphics[width=0.95\linewidth]{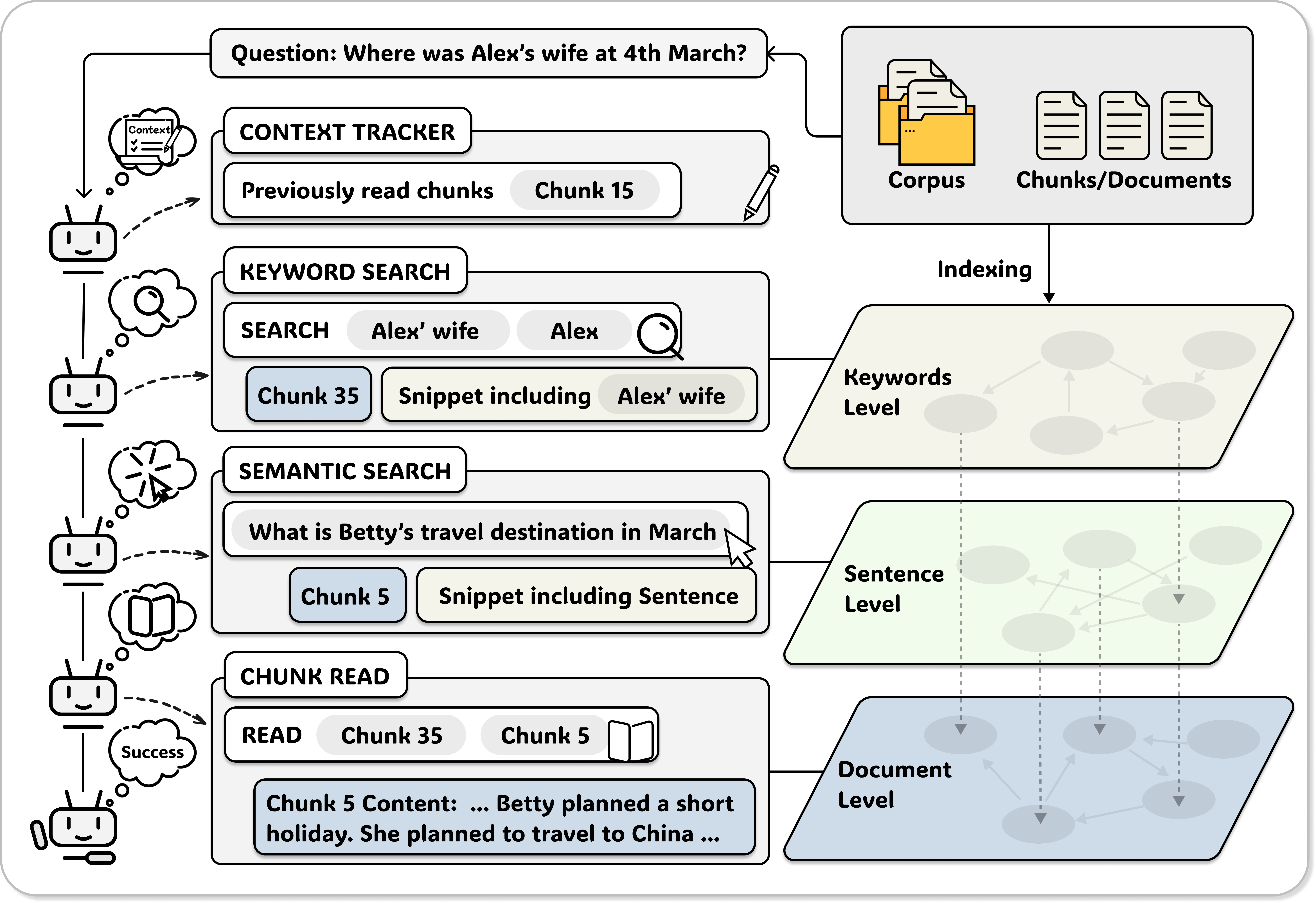}
  \caption{Overview of A-RAG framework. The agent iteratively uses hierarchical retrieval tools (keyword search, semantic search, chunk read) to gather information from the corpus and autonomously decides when to provide the final answer.}
  \label{fig:framework}
\end{figure}

\section{Methodology}

In this section, we present A-RAG, an agentic-RAG framework that exposes hierarchical retrieval interfaces to models. As illustrated in Figure~\ref{fig:framework}, our approach consists of three key components: (i) a hierarchical index, (ii) a suite of retrieval tools, and (iii) a simple agent loop design to clearly demonstrate the effectiveness of A-RAG.

\subsection{Hierarchical Index Construction}

To enable efficient multi-granularity retrieval, we construct a hierarchical index that organizes corpus information at different levels of abstraction. Our indexing procedure is lightweight and consists of only two stages: chunking and embedding.

\paragraph{Chunking.} Following the setup of LinearRAG~\citep{zhuang2025linearrag}, we partition the corpus into chunks of approximately 1,000 tokens each, ensuring that chunk boundaries align with sentence boundaries to preserve semantic coherence. Each chunk serves as a self-contained semantic unit that the agent can selectively access through dedicated retrieval interfaces, rather than being indiscriminately concatenated into the context as in conventional RAG approaches.

\paragraph{Embedding.} For each chunk $c_i$, we decompose it into sentences $\{s_{i,1}, s_{i,2}, \ldots, s_{i,n_i}\}$ using rule-based sentence segmentation. We then compute dense vector representations using a pre-trained sentence encoder $f_{\text{emb}}$: $\mathbf{v}_{i,j} = f_{\text{emb}}(s_{i,j})$. This sentence-level embedding enables fine-grained semantic matching while maintaining a mapping from sentences back to their parent chunks, allowing the agent to first identify relevant sentences and then read the complete chunk contexts.

\paragraph{Keyword-Level.} For the keyword-level information, we avoid pre-indexing. Instead of constructing inverted indices or knowledge graphs during the offline phase, we perform exact text matching directly at query time. This design choice significantly reduces both indexing time and computational cost compared to graph-based approaches. Through this lightweight indexing procedure, we obtain a three-level information representation: an implicit keyword-level for precise entity matching via runtime text search, sentence-level embeddings for semantic search, and chunk-level storage for full content access, which collectively support the hierarchical retrieval interfaces.

\subsection{Hierarchical Retrieval Interfaces}

We design three retrieval tools that operate at different granularities, enabling the agent to adaptively choose the most suitable search strategy based on the characteristics of each question.

\paragraph{Keyword Search.} This tool performs exact lexical matching to locate chunks containing specific terms. The agent provides a keyword list $\mathcal{K} = \{k_1, k_2, \ldots, k_m\}$ and a parameter $k$ specifying the number of results to return. The relevance score of chunk $c_i$ is computed as:
\begin{equation}
\text{Score}_{\text{kw}}(c_i, \mathcal{K}) = \sum_{k \in \mathcal{K}} \text{count}(k, T_i) \cdot |k|
\end{equation}
where $\text{count}(k, T_i)$ denotes the frequency of keyword $k$ in chunk text $T_i$, and $|k|$ is the character length of the keyword (longer keywords are weighted higher as they are typically more specific). For each matched chunk, we construct an abbreviated snippet by extracting sentences that contain at least one keyword:
\begin{equation}
\text{Snippet}(c_i, \mathcal{K}) = \{s \in \text{Sent}(c_i) \mid \exists k \in \mathcal{K}, k \subseteq s\}
\end{equation}
where $\text{Sent}(c_i)$ denotes the set of sentences in chunk $c_i$. The tool returns the top-$k$ chunk IDs along with their snippets, allowing the agent to autonomously decide the next action.

\paragraph{Semantic Search.} This tool finds semantically similar passages using dense retrieval. Given a natural language query $q$, we encode it into a query embedding $\mathbf{v}_q = f_{\text{emb}}(q)$ and compute cosine similarity with all sentence embeddings:
\begin{equation}
\text{Score}_{\text{sem}}(s_{i,j}, q) = \frac{\mathbf{v}_{i,j}^T \mathbf{v}_q}{\|\mathbf{v}_{i,j}\| \|\mathbf{v}_q\|}
\end{equation}
We retrieve the top-ranked sentences and aggregate them by their parent chunks. Each chunk's relevance score is determined by its highest-scoring sentence. The tool returns the top-$k$ chunk IDs along with the matched sentences within each chunk as snippets, allowing the agent to autonomously decide the next action.

\paragraph{Chunk Read.} Based on the snippets returned by keyword search and semantic search, the agent can determine which chunks require full reading and use this tool to access their complete content. The agent can also read adjacent chunks to gather additional context when needed.

This hierarchical design is inherently agent-friendly, allowing the agent to access corpus information at different granularities based on its own judgment. Rather than loading large amounts of context indiscriminately, the agent can incrementally retrieve information on-demand, minimizing context overhead while maintaining the flexibility to gather comprehensive evidence when needed.

\subsection{Agent Loop}

Since our method primarily focuses on interface design and investigating test-time scaling behavior in A-RAG, we deliberately adopt the simplest agent loop backbone to minimize confounding factors from complex orchestration mechanisms.

\paragraph{Agent Loop.} We adopt the ReAct-like framework~\citep{yao2023reactsynergizingreasoningacting}, where the model iteratively performs reasoning and tool calling in an interleaved manner. At each iteration, the agent selects \emph{one} tool to call, observes the result, and decides the next action. We intentionally avoid parallel tool calling and other sophisticated designs to facilitate clean observation of how different interface configurations influence agent behavior. When the maximum iteration budget is reached without producing an answer, we prompt the agent to synthesize a response based on the information gathered so far.

\paragraph{Context Tracker.} To prevent redundant information retrieval and unnecessary token consumption, we maintain a context tracker that records which chunks have been read during the retrieval process. Specifically, we track a set $\mathcal{C}^{\text{read}} = \{c_{i_1}, c_{i_2}, \ldots, c_{i_k}\}$, where each $c_{i_j}$ denotes the ID of a previously accessed chunk. When the agent attempts to read a chunk $c_i \in \mathcal{C}^{\text{read}}$, instead of returning the full text again, the chunk read tool returns a notification message ``This chunk has been read before'', consuming zero additional tokens. This mechanism not only reduces computational cost but also encourages the agent to explore diverse parts of the corpus rather than repeatedly examining the same passages.

This straightforward design allows us to cleanly isolate and analyze the impact of hierarchical interfaces on agent behavior and retrieval performance.

\section{Experiments}

In this section, we conduct comprehensive experiments to evaluate the effectiveness of A-RAG across multiple benchmarks and analyze its test-time scaling behavior.

\subsection{Experimental Setting}

\paragraph{Datasets.} We evaluate A-RAG on four widely-used multi-hop QA datasets: HotpotQA~\citep{yang2018hotpotqadatasetdiverseexplainable}, 2WikiMultiHopQA~\citep{ho2020constructingmultihopqadataset}, MuSiQue~\citep{trivedi2022musiquemultihopquestionssinglehop}, and GraphRAG-Bench~\citep{xiang2025usegraphsragcomprehensive}. Following the experimental setup of LinearRAG~\citep{zhuang2025linearrag}, we use the same corpus and questions to ensure fair comparison across different methods.

\begin{table}[!htbp]
  \centering
  \caption{Results (\%) of baselines and A-RAG on benchmark datasets in terms of LLM-Evaluation Accuracy(LLM-Acc) and Contain-Match Accuracy(Cont-Acc). The best result for each backbone LLM is highlighted in \textbf{bold}, while the second result is indicated with an \underline{underline}.}
  \label{tab:main-results}
  \small
  \renewcommand{\arraystretch}{1.15}
  \begin{tabular*}{\linewidth}{@{\extracolsep{\fill}}lcccccccc}
  \toprule
  \multirow{2}{*}{\textbf{Method}} & \multicolumn{2}{c}{\textbf{MuSiQue}} & \multicolumn{2}{c}{\textbf{HotpotQA}} & \multicolumn{2}{c}{\textbf{2Wiki}} & \textbf{Med.} & \textbf{Novel} \\
  \cmidrule(lr){2-3} \cmidrule(lr){4-5} \cmidrule(lr){6-7} \cmidrule(lr){8-8} \cmidrule(lr){9-9}
  & LLM & Cont & LLM & Cont & LLM & Cont & LLM & LLM \\
  \midrule
  \multicolumn{9}{c}{\textbf{\textit{GPT-4o-mini}}} \\
  \midrule
  \rowcolor{vanillacolor}
  \multicolumn{9}{c}{\textit{Vanilla Baselines}} \\
  Direct Answer & 18.3 & 13.9 & 45.4 & 40.7 & 30.3 & 49.7 & 68.6 & 45.3 \\
  Naive RAG & 38.6 & 36.1 & 74.5 & \underline{72.9} & 42.6 & 59.0 & 75.3 & 68.5 \\
  \midrule
  \rowcolor{graphworkflowcolor}
  \multicolumn{9}{c}{\textit{Graph-RAG and Workflow RAG}} \\
  GraphRAG & 26.4 & 20.8 & 33.2 & 33.3 & 18.4 & 47.2 & 51.3 & 28.8 \\
  HippoRAG2 & 40.6 & 38.4 & \textbf{80.7} & 69.7 & \textbf{64.7} & \textbf{68.5} & 72.0 & \underline{70.1} \\
  LinearRAG & 34.8 & 26.3 & 72.0 & 60.5 & \underline{62.9} & \underline{62.3} & 53.1 & 45.4 \\
  FaithfulRAG & 28.8 & 22.6 & 60.5 & 52.5 & 38.8 & 38.1 & 42.5 & 33.3 \\
  MA-RAG & 34.1 & 27.4 & 60.6 & 54.4 & 51.0 & 53.4 & 62.3 & 44.5 \\
  RAGentA & 32.2 & 29.9 & 63.0 & 62.4 & 27.7 & 50.3 & 67.7 & 61.3 \\
  \midrule
  \rowcolor{aragcolor}
  \multicolumn{9}{c}{\textit{A-RAG (Ours)}} \\
  \textbf{A-RAG (Naive)} & \underline{43.8} & \underline{38.5} & 76.6 & 70.7 & 52.3 & 62.4 & \underline{79.0} & 70.0 \\
  \textbf{A-RAG (Full)} & \textbf{46.1} & \textbf{39.6} & \underline{77.1} & \textbf{74.0} & 60.2 & 63.7 & \textbf{79.4} & \textbf{72.7} \\
  \midrule
  \multicolumn{9}{c}{\textbf{\textit{GPT-5-mini}}} \\
  \midrule
  \rowcolor{vanillacolor}
  \multicolumn{9}{c}{\textit{Vanilla Baselines}} \\
  Direct Answer & 35.8 & 26.5 & 63.6 & 53.5 & 51.3 & 54.0 & 90.5 & 45.1 \\
  Naive RAG & 52.8 & 48.7 & 81.2 & 79.5 & 50.2 & 66.5 & 86.1 & 70.6 \\
  \midrule
  \rowcolor{graphworkflowcolor}
  \multicolumn{9}{c}{\textit{Graph-RAG and Workflow RAG}} \\
  GraphRAG & 48.3 & 39.1 & 82.5 & 74.9 & 66.5 & 70.7 & 87.3 & \underline{77.1} \\
  HippoRAG2 & 61.7 & 52.5 & 84.8 & 75.0 & 82.0 & 79.7 & 78.2 & 54.3 \\
  LinearRAG & 62.4 & 51.8 & 86.2 & 77.6 & \underline{87.2} & \underline{84.8} & 79.2 & 54.7 \\
  FaithfulRAG & 52.9 & 52.8 & 76.9 & 75.3 & 51.8 & 56.6 & 75.4 & 60.7 \\
  MA-RAG & 40.0 & 31.6 & 67.1 & 57.9 & 54.7 & 54.3 & 68.3 & 45.1 \\
  RAGentA & 38.3 & 37.4 & 61.2 & 65.0 & 24.0 & 53.5 & 73.7 & 60.2 \\
  \midrule
  \rowcolor{aragcolor}
  \multicolumn{9}{c}{\textit{A-RAG (Ours)}} \\
  \textbf{A-RAG (Naive)} & \underline{66.2} & \underline{59.7} & \underline{90.8} & \underline{85.3} & 70.6 & 76.9 & \underline{92.7} & 80.4 \\
  \textbf{A-RAG (Full)} & \textbf{74.1} & \textbf{65.3} & \textbf{94.5} & \textbf{88.0} & \textbf{89.7} & \textbf{88.9} & \textbf{93.1} & \textbf{85.3} \\
  \bottomrule
  \end{tabular*}
\end{table}

\paragraph{Baselines.} We organize all compared methods into two groups: (i) \textbf{Vanilla Baselines}: including direct zero-shot LLM inference and Naive RAG method; (ii) \textbf{Graph-RAG and Workflow RAG}: we benchmark against representative graph-enhanced approaches including GraphRAG~\citep{edge2025localglobalgraphrag}, HippoRAG2~\citep{gutiérrez2025hipporagneurobiologicallyinspiredlongterm}, and LinearRAG~\citep{zhuang2025linearrag}, as well as workflow-based methods including FaithfulRAG~\citep{zhang2025faithfulragfactlevelconflictmodeling}, MA-RAG~\citep{nguyen2025maragmultiagentretrievalaugmentedgeneration}, and RAGentA~\citep{besrour2025ragentamultiagentretrievalaugmentedgeneration,chang2024mainragmultiagentfilteringretrievalaugmented}. We compare these baselines against our A-RAG (Naive), equipped with only a single embedding search tool, and A-RAG (Full).

\paragraph{Evaluation Metrics.} Following LinearRAG, we employ two metrics for end-to-end QA assessment: (1) LLM-Evaluation Accuracy (LLM-Acc, corresponding to GPT-Acc in LinearRAG), an LLM-based metric that determines semantic equivalence between predictions and ground-truth answers, and (2) Contain-Match Accuracy (Contain-Acc), which verifies whether the ground-truth answer appears within the generated response. For HotpotQA, 2WikiMultiHopQA, and MuSiQue with short-form answers, we report both metrics. For GraphRAG-Bench with long-form descriptive answers, we report LLM-Acc only, as lengthy ground-truth answers rarely appear verbatim in generated responses, making Contain-Acc uninformative.

\paragraph{Implementation.} We evaluate all methods using both GPT-4o-mini and GPT-5-mini~\citep{openai_gpt5_system_card_2025} as backbone LLMs. For dense retrieval, all methods except LinearRAG utilize Qwen3-Embedding-0.6B~\citep{zhang2025qwen3embeddingadvancingtext} with $k=5$ for top-$k$ results; LinearRAG uses its original embedding model due to incompatibility between its NER module and Qwen3-Embedding. We intentionally include both earlier and frontier reasoning models to provide a comprehensive view of how RAG methods perform across different capability levels. For LLM-based evaluation, we use GPT-5-mini as the judge, which demonstrates improved accuracy and stability based on our human verification. Detailed configuration and hyperparameters are provided in Appendix~\ref{sec:baseline-details}.

\begin{table}[!htbp]
  \centering
  \caption{Ablation study results (\%) on benchmark datasets.}
  \label{tab:ablation}
  \small
  \renewcommand{\arraystretch}{1.15}
  \begin{tabular*}{\linewidth}{@{\extracolsep{\fill}}lcccccccc}
  \toprule
  \multirow{2}{*}{\textbf{Method}} & \multicolumn{2}{c}{\textbf{MuSiQue}} & \multicolumn{2}{c}{\textbf{HotpotQA}} & \multicolumn{2}{c}{\textbf{2Wiki}} & \textbf{Med.} & \textbf{Novel} \\
  \cmidrule(lr){2-3} \cmidrule(lr){4-5} \cmidrule(lr){6-7} \cmidrule(lr){8-8} \cmidrule(lr){9-9}
  & LLM & Cont & LLM & Cont & LLM & Cont & LLM & LLM \\
  \midrule
  A-RAG (Full) & \cellcolor{aragcolor}\textbf{74.1} & 65.3 & \cellcolor{aragcolor}\textbf{94.5} & 88.0 & \cellcolor{aragcolor}\textbf{89.7} & \cellcolor{aragcolor}\textbf{88.9} & 93.1 & \cellcolor{aragcolor}\textbf{85.3} \\
  \quad w/o KW Search & 72.6 & 65.3 & 93.0 & 87.4 & 88.9 & 88.1 & 93.2 & 85.0 \\
  \quad w/o Semantic & 69.4 & 63.3 & 93.9 & 88.4 & 89.1 & 88.0 & 92.1 & 85.2 \\
  \quad w/o Chunk Read & 73.6 & \cellcolor{aragcolor}\textbf{67.0} & 93.6 & \cellcolor{aragcolor}\textbf{88.8} & 89.0 & 87.9 & \cellcolor{aragcolor}\textbf{93.3} & 85.1 \\
  \bottomrule
  \end{tabular*}
\end{table}

\subsection{Main Results}

Table~\ref{tab:main-results} presents the main experimental results across all benchmarks. We highlight three key observations from our experiments:

\paragraph{Vanilla retrieval method remain robust baseline.} Under our unified evaluation setting with GPT-5-mini as the judge and Qwen3-Embedding for dense retrieval, vanilla baselines demonstrate robust performance across both GPT-4o-mini and GPT-5-mini backbones. Existing Graph-RAG and Workflow RAG methods fail to consistently outperform these simple baselines across all datasets.

\paragraph{Naive A-RAG establishes a new strong baseline for agentic RAG.} As a simplified variant equipped with only a single embedding-based retrieval tool, A-RAG (Naive) surpasses existing Graph-RAG and Workflow RAG methods on multiple datasets, demonstrating the inherent advantages of the agentic paradigm. This advantage becomes more pronounced when switching to GPT-5-mini as the backbone. This result suggests that granting models greater autonomy in retrieval decisions yields better performance than relying on fixed retrieval algorithms, even without sophisticated multi-granularity tools.

\paragraph{A-RAG outperforms existing RAG methods through hierarchical retrieval interfaces.} A-RAG is designed for reasoning models with tool-use capabilities, aligning with the current development trend of the LLM field. With GPT-4o-mini as the backbone, A-RAG (Full) achieves the best performance on 3 out of 5 datasets. When switching to GPT-5-mini with stronger reasoning and tool-calling capabilities, A-RAG (Full) achieves superior results across all benchmarks. The consistent improvements of A-RAG over both baseline methods and Naive A-RAG demonstrate that the A-RAG framework is agent-friendly. It allows models to leverage their reasoning capabilities to dynamically adjust strategies and orchestrate different interfaces based on task requirements, thereby achieving better performance.

\subsection{Ablation Study}

To investigate the contribution of each retrieval tool, we conduct ablation experiments by systematically removing individual components from A-RAG (Full). We evaluate three ablation variants: (i) w/o Keyword Search and w/o Semantic Search, which directly remove the corresponding retrieval tool from the agent's toolkit; (ii) w/o Chunk Read, which replaces the snippet-based responses of keyword and semantic search with complete chunk texts and removes the chunk read tool entirely.

\textbf{As shown in Table~\ref{tab:ablation}, the full hierarchical configuration achieves optimal overall performance.} A-RAG (Full) consistently achieves the best results on most benchmarks. Removing either semantic search or keyword search leads to performance degradation, highlighting the importance of multi-granularity information for multi-hop retrieval tasks. The inferior performance of w/o Chunk Read compared to A-RAG (Full) demonstrates that our progressive information acquisition design allows the agent to make autonomous judgments and precisely read the most relevant content. This design not only enhances agent autonomy but also enables the model to selectively read only the most relevant chunks in full, avoiding the noise introduced by irrelevant content.

\begin{figure}[!htbp]
  \centering
  \includegraphics[width=0.95\linewidth]{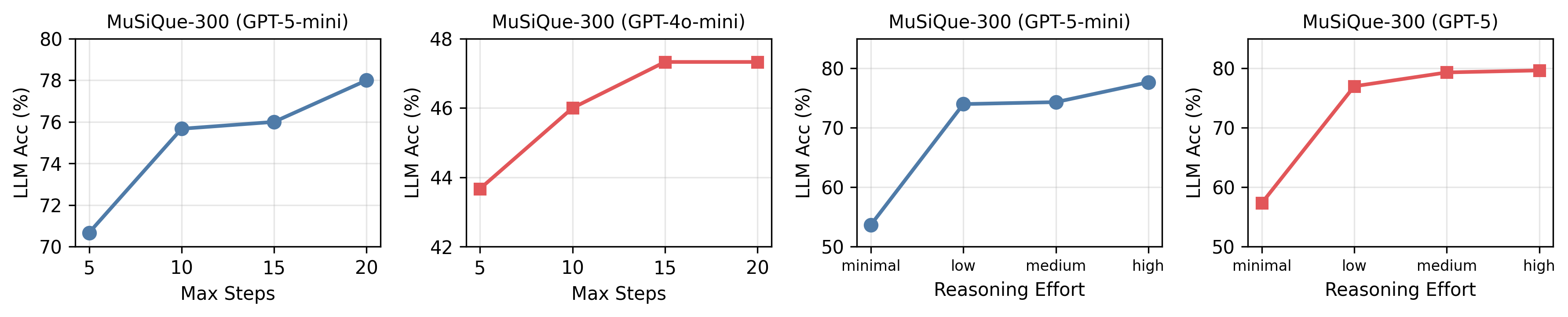}
  \caption{Test-time scaling analysis on MuSiQue-300. Left two: LLM-Acc vs. max steps with GPT-5-mini and GPT-4o-mini. Right two: LLM-Acc vs. reasoning effort with GPT-5-mini and GPT-5.}
  \label{fig:depth-scaling}
\end{figure}

\section{Analysis and Discussion}

To understand the advantages and characteristics of A-RAG as a new paradigm, we conduct further experiments and analyses in this section.

\subsection{Test-Time Scaling Analysis}

Since A-RAG grants LLMs greater autonomy in retrieval decisions, increasing computational resources at test time can further scale the framework's performance. As shown in Figure~\ref{fig:depth-scaling}, we conduct experiments on the first 300 tasks of MuSiQue and find that both increasing max-step and reasoning effort effectively scale model performance. When scaling from 5 to 20 steps, GPT-5-mini improves by approximately 8\% while GPT-4o-mini improves by only about 4\%, indicating that stronger reasoning models are better equipped for longer-horizon exploration. When scaling reasoning effort from minimal to high, both GPT-5-mini and GPT-5 achieve substantial improvements of approximately 25\%. These results demonstrate that A-RAG effectively leverages test-time compute, positioning it as a promising paradigm for future development.

\subsection{Context Efficiency Analysis}

Context efficiency is crucial for integrating RAG into complex agentic systems. We analyze the tokens retrieved from the corpus to measure how efficiently each method utilizes context (Table~\ref{tab:retrieval-tokens}).

\begin{table}[!htbp]
  \centering
  \caption{Retrieved tokens across methods (GPT-5-mini backbone). Lower values indicate higher efficiency.}
  \label{tab:retrieval-tokens}
  \small
  \renewcommand{\arraystretch}{1.15}
  \begin{tabular*}{\linewidth}{@{\extracolsep{\fill}}lccccc}
  \toprule
  \textbf{Method} & \textbf{MuSi.} & \textbf{Hotpot.} & \textbf{2Wiki} & \textbf{Med.} & \textbf{Novel} \\
  \midrule
  Naive RAG & 5,387 & 5,358 & 5,506 & 5,418 & 4,997 \\
  HippoRAG2 & 5,411 & 5,380 & 5,538 & 5,447 & 5,019 \\
  GraphRAG & 9,234 & 8,744 & 4,201 & 9,391 & 9,318 \\
  LinearRAG & 5,418 & 5,353 & 5,518 & 5,427 & 4,998 \\
  FaithfulRAG & \textbf{5,342} & 5,310 & 5,419 & \textbf{5,410} & \textbf{4,994} \\
  MA-RAG & 9,566 & 8,007 & 8,857 & 6,858 & 6,101 \\
  \midrule
  A-RAG (Naive) & 56,360 & 27,455 & 45,406 & 23,657 & 22,391 \\
  A-RAG (Full) & 5,663 & \textbf{2,737} & \textbf{2,930} & 7,678 & 6,087 \\
  \bottomrule
  \end{tabular*}
\end{table}

\textbf{A-RAG achieves superior accuracy with higher context efficiency.} Contrary to the intuition that more retrieved content leads to better performance, A-RAG (Full) retrieves comparable or fewer tokens than traditional RAG methods while achieving superior accuracy.

\begin{wrapfigure}{r}{0.6\textwidth}
  \centering
  \vspace{-8pt}
  \includegraphics[width=0.98\linewidth]{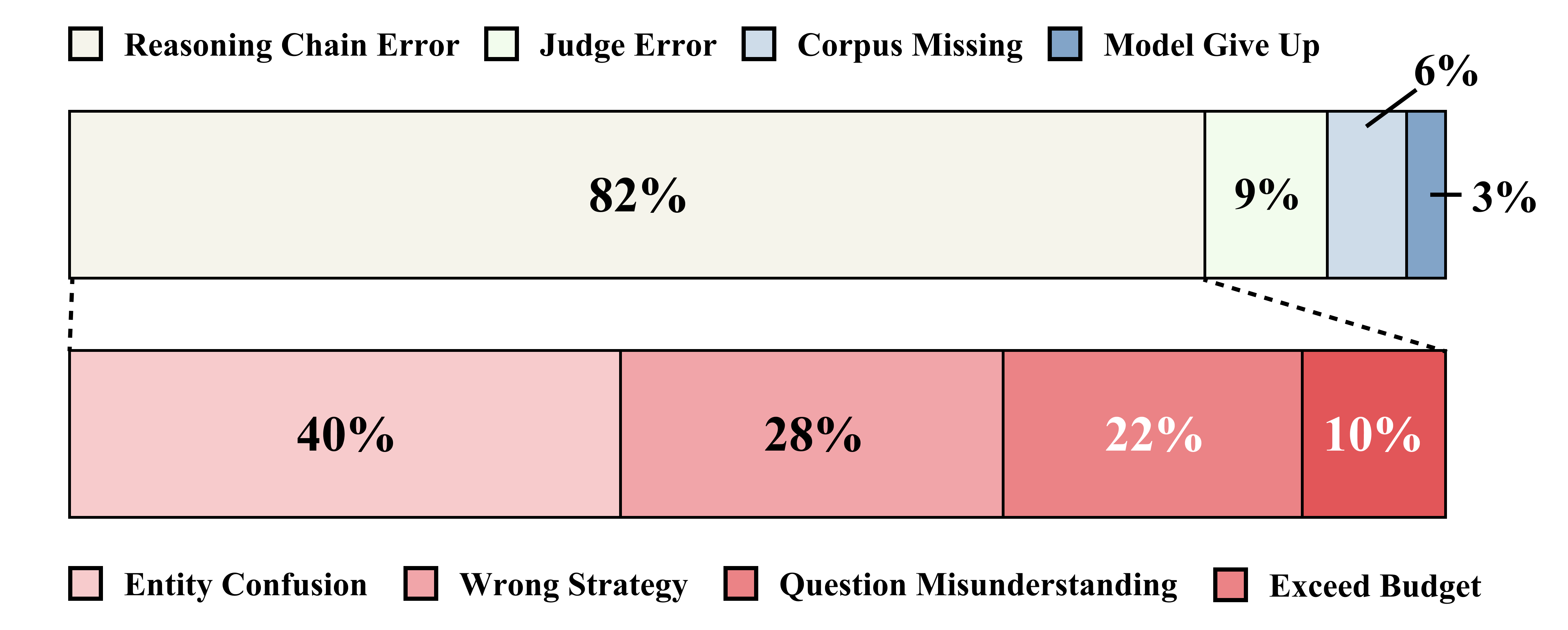}
  \caption{Failure mode distribution of A-RAG. Top: primary categories. Bottom: breakdown of reasoning chain errors.}
  \label{fig:failure-mode}
  \vspace{-8pt}
\end{wrapfigure}

\textbf{Hierarchical interfaces are key to context efficiency.} Comparing A-RAG (Naive) and A-RAG (Full) reveals a striking pattern: A-RAG (Naive) retrieves more tokens than A-RAG (Full) but achieves lower performance. This validates our hierarchical interface design, the progressive information disclosure grants the model greater autonomy while avoiding irrelevant content.

\subsection{Failure Mode Analysis}

We manually reviewed the first 100 incorrect cases of A-RAG on MuSiQue and categorized them into 2-level error types (Figure~\ref{fig:failure-mode}). The majority of failures stem from reasoning chain errors. Among these, entity confusion is the most common, with substantial portions also attributed to wrong retrieval strategies and question misunderstanding. Detailed category definitions and analysis on other datasets are provided in Appendix~\ref{sec:failure-details}.

\section{Conclusion}

In this work, we recognize agentic RAG as a fundamental paradigm shift in RAG. We introduce A-RAG, an agentic RAG framework featuring hierarchical retrieval interfaces that enable LLMs to autonomously access corpus information at keyword, sentence, and chunk levels. Extensive experiments demonstrate that A-RAG consistently outperforms existing Graph-RAG and Workflow RAG methods across diverse benchmarks, while our analysis validates its efficient test-time scaling behavior. Our findings suggest that future research should focus on designing agent-friendly interfaces rather than complex retrieval algorithms, and explore new interaction paradigms between language models and external knowledge sources.

\section*{Limitations}

Our work primarily aims to highlight the paradigm shift from traditional RAG to agentic RAG and demonstrate hierarchical interfaces as a promising scaling direction. However, we do not exhaustively enumerate all possible tool designs or systematically compare different tool subsets and their impacts on agent behavior. A comprehensive ablation across diverse tool configurations could provide deeper insights into optimal interface design, which we leave for future work.

Due to computational resource constraints, we have not validated the framework on larger and more powerful models such as GPT-5, and Gemini-3. Given that A-RAG is specifically designed for reasoning models with strong tool-use capabilities, we anticipate that performance gains would be more pronounced with these frontier models, but empirical verification remains to be conducted.

Additionally, while we demonstrate strong results on multi-hop QA benchmarks, the generalization of A-RAG to other knowledge-intensive tasks such as fact verification, dialogue systems, and long-form generation warrants further investigation.

\section*{Ethical Considerations}

All datasets used in this work are publicly available benchmarks that have been previously curated and processed by prior research with appropriate ethical considerations. Our work focuses on fundamental research for improving retrieval-augmented generation in large language models, and does not involve the collection of new data or human subjects. As a methodological contribution to RAG systems, our approach does not introduce additional ethical risks beyond those inherent to the underlying language models.

\newpage
\bibliography{custom}

@online{openai2025gpt5,
  author  = {{OpenAI}},
  title   = {Introducing GPT-5},
  year    = {2025},
  url     = {https://openai.com/zh-Hans-CN/index/introducing-gpt-5/},
  urldate = {2025-12-04}
}

@online{anthropic2025claude45sonnet,
  author  = {{Anthropic}},
  title   = {Introducing Claude Sonnet 4.5},
  year    = {2025},
  url     = {https://www.anthropic.com/news/claude-sonnet-4-5},
  urldate = {2025-12-04}
}

@online{google2025gemini3,
  author  = {{Google}},
  title   = {A New Era of Intelligence with Gemini 3},
  year    = {2025},
  url     = {https://blog.google/products/gemini/gemini-3/},
  urldate = {2025-12-04}
}

@misc{deepseekai2025deepseekr1incentivizingreasoningcapability,
  title         = {DeepSeek-R1: Incentivizing Reasoning Capability in LLMs via Reinforcement Learning},
  author        = {DeepSeek-AI and Daya Guo and Dejian Yang and Haowei Zhang and Junxiao Song and Ruoyu Zhang and Runxin Xu and others},
  year          = {2025},
  eprint        = {2501.12948},
  archivePrefix = {arXiv},
  primaryClass  = {cs.CL},
  url           = {https://arxiv.org/abs/2501.12948}
}

@misc{shao2025deepseekmathv2selfverifiablemathematicalreasoning,
  title         = {DeepSeekMath-V2: Towards Self-Verifiable Mathematical Reasoning},
  author        = {Zhihong Shao and Yuxiang Luo and Chengda Lu and Z. Z. Ren and Jiewen Hu and Tian Ye and Zhibin Gou and Shirong Ma and Xiaokang Zhang},
  year          = {2025},
  eprint        = {2511.22570},
  archivePrefix = {arXiv},
  primaryClass  = {cs.AI},
  url           = {https://arxiv.org/abs/2511.22570}
}

@misc{yang2025qwen3technicalreport,
  title         = {Qwen3 Technical Report},
  author        = {An Yang and Anfeng Li and Baosong Yang and Beichen Zhang and Binyuan Hui and Bo Zheng and others},
  year          = {2025},
  eprint        = {2505.09388},
  archivePrefix = {arXiv},
  primaryClass  = {cs.CL},
  url           = {https://arxiv.org/abs/2505.09388}
}

@misc{kimiteam2025kimik2openagentic,
  title         = {Kimi K2: Open Agentic Intelligence},
  author        = {Kimi Team and Yifan Bai and Yiping Bao and Guanduo Chen and Jiahao Chen and Ningxin Chen and others},
  year          = {2025},
  eprint        = {2507.20534},
  archivePrefix = {arXiv},
  primaryClass  = {cs.LG},
  url           = {https://arxiv.org/abs/2507.20534}
}

@online{minimax2025m2,
  author  = {{MiniMax AI}},
  title   = {MiniMax-M2: A Model Built for Max Coding \& Agentic Workflows},
  year    = {2025},
  url     = {https://github.com/MiniMax-AI/MiniMax-M2},
  urldate = {2025-12-04}
}

@misc{cursor_ide,
  author  = {{Anysphere, Inc.}},
  title   = {Cursor: AI Code Editor},
  year    = {2023},
  howpublished = {\url{https://www.cursor.com/}},
  note = {Accessed: 2025-12-04}
}

@misc{claude_code,
  author  = {{Anthropic}},
  title   = {Claude Code: Agentic Coding Assistant},
  year    = {2025},
  howpublished = {\url{https://code.claude.com/docs/en/overview}},
  note = {Accessed: 2025-12-04}
}

@misc{chatgpt_deep_research,
  author  = {{OpenAI}},
  title   = {Introducing Deep Research},
  year    = {2025},
  howpublished = {\url{https://openai.com/index/introducing-deep-research/}},
  note = {Accessed: 2025-12-04}
}

@misc{gemini_deep_research,
  author  = {{Google LLC}},
  title   = {Gemini Deep Research: Your Personal Research Assistant},
  year    = {2024},
  howpublished = {\url{https://gemini.google/overview/deep-research/}},
  note = {Accessed: 2025-12-04}
}

@misc{tongyideepresearchteam2025tongyideepresearchtechnicalreport,
  title         = {Tongyi DeepResearch Technical Report},
  author        = {{Tongyi DeepResearch Team}},
  year          = {2025},
  eprint        = {2510.24701},
  archivePrefix = {arXiv},
  primaryClass  = {cs.CL},
  url           = {https://arxiv.org/abs/2510.24701}
}

@misc{lewis2021retrievalaugmentedgenerationknowledgeintensivenlp,
      title={Retrieval-Augmented Generation for Knowledge-Intensive NLP Tasks}, 
      author={Patrick Lewis and Ethan Perez and Aleksandra Piktus and Fabio Petroni and Vladimir Karpukhin and Naman Goyal and Heinrich Küttler and Mike Lewis and Wen-tau Yih and Tim Rocktäschel and Sebastian Riedel and Douwe Kiela},
      year={2021},
      eprint={2005.11401},
      archivePrefix={arXiv},
      primaryClass={cs.CL},
      url={https://arxiv.org/abs/2005.11401}, 
}

@misc{asai2023selfraglearningretrievegenerate,
      title={Self-RAG: Learning to Retrieve, Generate, and Critique through Self-Reflection}, 
      author={Akari Asai and Zeqiu Wu and Yizhong Wang and Avirup Sil and Hannaneh Hajishirzi},
      year={2023},
      eprint={2310.11511},
      archivePrefix={arXiv},
      primaryClass={cs.CL},
      url={https://arxiv.org/abs/2310.11511}, 
}

@misc{yan2024correctiveretrievalaugmentedgeneration,
      title={Corrective Retrieval Augmented Generation}, 
      author={Shi-Qi Yan and Jia-Chen Gu and Yun Zhu and Zhen-Hua Ling},
      year={2024},
      eprint={2401.15884},
      archivePrefix={arXiv},
      primaryClass={cs.CL},
      url={https://arxiv.org/abs/2401.15884}, 
}

@misc{jeong2024adaptiveraglearningadaptretrievalaugmented,
      title={Adaptive-RAG: Learning to Adapt Retrieval-Augmented Large Language Models through Question Complexity}, 
      author={Soyeong Jeong and Jinheon Baek and Sukmin Cho and Sung Ju Hwang and Jong C. Park},
      year={2024},
      eprint={2403.14403},
      archivePrefix={arXiv},
      primaryClass={cs.CL},
      url={https://arxiv.org/abs/2403.14403}, 
}

@misc{jiang2023activeretrievalaugmentedgeneration,
      title={Active Retrieval Augmented Generation}, 
      author={Zhengbao Jiang and Frank F. Xu and Luyu Gao and Zhiqing Sun and Qian Liu and Jane Dwivedi-Yu and Yiming Yang and Jamie Callan and Graham Neubig},
      year={2023},
      eprint={2305.06983},
      archivePrefix={arXiv},
      primaryClass={cs.CL},
      url={https://arxiv.org/abs/2305.06983}, 
}

@misc{trivedi2023interleavingretrievalchainofthoughtreasoning,
      title={Interleaving Retrieval with Chain-of-Thought Reasoning for Knowledge-Intensive Multi-Step Questions}, 
      author={Harsh Trivedi and Niranjan Balasubramanian and Tushar Khot and Ashish Sabharwal},
      year={2023},
      eprint={2212.10509},
      archivePrefix={arXiv},
      primaryClass={cs.CL},
      url={https://arxiv.org/abs/2212.10509}, 
}

@misc{chan2024rqraglearningrefinequeries,
      title={RQ-RAG: Learning to Refine Queries for Retrieval Augmented Generation}, 
      author={Chi-Min Chan and Chunpu Xu and Ruibin Yuan and Hongyin Luo and Wei Xue and Yike Guo and Jie Fu},
      year={2024},
      eprint={2404.00610},
      archivePrefix={arXiv},
      primaryClass={cs.CL},
      url={https://arxiv.org/abs/2404.00610}, 
}

@misc{liu2024raisflearninganswerunderstand,
      title={RA-ISF: Learning to Answer and Understand from Retrieval Augmentation via Iterative Self-Feedback}, 
      author={Yanming Liu and Xinyue Peng and Xuhong Zhang and Weihao Liu and Jianwei Yin and Jiannan Cao and Tianyu Du},
      year={2024},
      eprint={2403.06840},
      archivePrefix={arXiv},
      primaryClass={cs.CL},
      url={https://arxiv.org/abs/2403.06840}, 
}

@misc{guo2025lightragsimplefastretrievalaugmented,
      title={LightRAG: Simple and Fast Retrieval-Augmented Generation}, 
      author={Zirui Guo and Lianghao Xia and Yanhua Yu and Tu Ao and Chao Huang},
      year={2025},
      eprint={2410.05779},
      archivePrefix={arXiv},
      primaryClass={cs.IR},
      url={https://arxiv.org/abs/2410.05779}, 
}

@misc{qian2025memoragboostinglongcontext,
      title={MemoRAG: Boosting Long Context Processing with Global Memory-Enhanced Retrieval Augmentation}, 
      author={Hongjin Qian and Zheng Liu and Peitian Zhang and Kelong Mao and Defu Lian and Zhicheng Dou and Tiejun Huang},
      year={2025},
      eprint={2409.05591},
      archivePrefix={arXiv},
      primaryClass={cs.CL},
      url={https://arxiv.org/abs/2409.05591}, 
}

@misc{gutiérrez2025hipporagneurobiologicallyinspiredlongterm,
      title={HippoRAG: Neurobiologically Inspired Long-Term Memory for Large Language Models}, 
      author={Bernal Jiménez Gutiérrez and Yiheng Shu and Yu Gu and Michihiro Yasunaga and Yu Su},
      year={2025},
      eprint={2405.14831},
      archivePrefix={arXiv},
      primaryClass={cs.CL},
      url={https://arxiv.org/abs/2405.14831}, 
}

@misc{gutiérrez2025ragmemorynonparametriccontinual,
      title={From RAG to Memory: Non-Parametric Continual Learning for Large Language Models}, 
      author={Bernal Jiménez Gutiérrez and Yiheng Shu and Weijian Qi and Sizhe Zhou and Yu Su},
      year={2025},
      eprint={2502.14802},
      archivePrefix={arXiv},
      primaryClass={cs.CL},
      url={https://arxiv.org/abs/2502.14802}, 
}

@misc{sarthi2024raptorrecursiveabstractiveprocessing,
      title={RAPTOR: Recursive Abstractive Processing for Tree-Organized Retrieval}, 
      author={Parth Sarthi and Salman Abdullah and Aditi Tuli and Shubh Khanna and Anna Goldie and Christopher D. Manning},
      year={2024},
      eprint={2401.18059},
      archivePrefix={arXiv},
      primaryClass={cs.CL},
      url={https://arxiv.org/abs/2401.18059}, 
}

@misc{edge2025localglobalgraphrag,
      title={From Local to Global: A Graph RAG Approach to Query-Focused Summarization}, 
      author={Darren Edge and Ha Trinh and Newman Cheng and Joshua Bradley and Alex Chao and Apurva Mody and Steven Truitt and Dasha Metropolitansky and Robert Osazuwa Ness and Jonathan Larson},
      year={2025},
      eprint={2404.16130},
      archivePrefix={arXiv},
      primaryClass={cs.CL},
      url={https://arxiv.org/abs/2404.16130}, 
}

@misc{shen2025geargraphenhancedagentretrievalaugmented,
      title={GeAR: Graph-enhanced Agent for Retrieval-augmented Generation}, 
      author={Zhili Shen and Chenxin Diao and Pavlos Vougiouklis and Pascual Merita and Shriram Piramanayagam and Enting Chen and Damien Graux and Andre Melo and Ruofei Lai and Zeren Jiang and Zhongyang Li and YE QI and Yang Ren and Dandan Tu and Jeff Z. Pan},
      year={2025},
      eprint={2412.18431},
      archivePrefix={arXiv},
      primaryClass={cs.CL},
      url={https://arxiv.org/abs/2412.18431}, 
}

@misc{yang2025graphsearchagenticdeepsearching,
      title={GraphSearch: An Agentic Deep Searching Workflow for Graph Retrieval-Augmented Generation}, 
      author={Cehao Yang and Xiaojun Wu and Xueyuan Lin and Chengjin Xu and Xuhui Jiang and Yuanliang Sun and Jia Li and Hui Xiong and Jian Guo},
      year={2025},
      eprint={2509.22009},
      archivePrefix={arXiv},
      primaryClass={cs.CL},
      url={https://arxiv.org/abs/2509.22009}, 
}

@misc{song2025efficienttransferableagenticknowledge,
      title={Efficient and Transferable Agentic Knowledge Graph RAG via Reinforcement Learning}, 
      author={Jinyeop Song and Song Wang and Julian Shun and Yada Zhu},
      year={2025},
      eprint={2509.26383},
      archivePrefix={arXiv},
      primaryClass={cs.CL},
      url={https://arxiv.org/abs/2509.26383}, 
}

@misc{nguyen2025maragmultiagentretrievalaugmentedgeneration,
      title={MA-RAG: Multi-Agent Retrieval-Augmented Generation via Collaborative Chain-of-Thought Reasoning}, 
      author={Thang Nguyen and Peter Chin and Yu-Wing Tai},
      year={2025},
      eprint={2505.20096},
      archivePrefix={arXiv},
      primaryClass={cs.CL},
      url={https://arxiv.org/abs/2505.20096}, 
}

@misc{besrour2025ragentamultiagentretrievalaugmentedgeneration,
      title={RAGentA: Multi-Agent Retrieval-Augmented Generation for Attributed Question Answering}, 
      author={Ines Besrour and Jingbo He and Tobias Schreieder and Michael Färber},
      year={2025},
      eprint={2506.16988},
      archivePrefix={arXiv},
      primaryClass={cs.IR},
      url={https://arxiv.org/abs/2506.16988}, 
}

@misc{huang2025retrievalaugmentedgenerationhierarchicalknowledge,
      title={Retrieval-Augmented Generation with Hierarchical Knowledge}, 
      author={Haoyu Huang and Yongfeng Huang and Junjie Yang and Zhenyu Pan and Yongqiang Chen and Kaili Ma and Hongzhi Chen and James Cheng},
      year={2025},
      eprint={2503.10150},
      archivePrefix={arXiv},
      primaryClass={cs.CL},
      url={https://arxiv.org/abs/2503.10150}, 
}

@misc{chang2024mainragmultiagentfilteringretrievalaugmented,
      title={MAIN-RAG: Multi-Agent Filtering Retrieval-Augmented Generation}, 
      author={Chia-Yuan Chang and Zhimeng Jiang and Vineeth Rakesh and Menghai Pan and Chin-Chia Michael Yeh and Guanchu Wang and Mingzhi Hu and Zhichao Xu and Yan Zheng and Mahashweta Das and Na Zou},
      year={2024},
      eprint={2501.00332},
      archivePrefix={arXiv},
      primaryClass={cs.CL},
      url={https://arxiv.org/abs/2501.00332}, 
}

@misc{chen2025improvingretrievalaugmentedgenerationmultiagent,
      title={Improving Retrieval-Augmented Generation through Multi-Agent Reinforcement Learning}, 
      author={Yiqun Chen and Lingyong Yan and Weiwei Sun and Xinyu Ma and Yi Zhang and Shuaiqiang Wang and Dawei Yin and Yiming Yang and Jiaxin Mao},
      year={2025},
      eprint={2501.15228},
      archivePrefix={arXiv},
      primaryClass={cs.CL},
      url={https://arxiv.org/abs/2501.15228}, 
}

@misc{xiong2025raggymsystematicoptimizationlanguage,
      title={RAG-Gym: Systematic Optimization of Language Agents for Retrieval-Augmented Generation}, 
      author={Guangzhi Xiong and Qiao Jin and Xiao Wang and Yin Fang and Haolin Liu and Yifan Yang and Fangyuan Chen and Zhixing Song and Dengyu Wang and Minjia Zhang and Zhiyong Lu and Aidong Zhang},
      year={2025},
      eprint={2502.13957},
      archivePrefix={arXiv},
      primaryClass={cs.CL},
      url={https://arxiv.org/abs/2502.13957}, 
}

@misc{jin2025searchr1trainingllmsreason,
      title={Search-R1: Training LLMs to Reason and Leverage Search Engines with Reinforcement Learning}, 
      author={Bowen Jin and Hansi Zeng and Zhenrui Yue and Jinsung Yoon and Sercan Arik and Dong Wang and Hamed Zamani and Jiawei Han},
      year={2025},
      eprint={2503.09516},
      archivePrefix={arXiv},
      primaryClass={cs.CL},
      url={https://arxiv.org/abs/2503.09516}, 
}

@misc{song2025r1searcherincentivizingsearchcapability,
      title={R1-Searcher: Incentivizing the Search Capability in LLMs via Reinforcement Learning}, 
      author={Huatong Song and Jinhao Jiang and Yingqian Min and Jie Chen and Zhipeng Chen and Wayne Xin Zhao and Lei Fang and Ji-Rong Wen},
      year={2025},
      eprint={2503.05592},
      archivePrefix={arXiv},
      primaryClass={cs.AI},
      url={https://arxiv.org/abs/2503.05592}, 
}

@misc{luo2025graphr1agenticgraphragframework,
      title={Graph-R1: Towards Agentic GraphRAG Framework via End-to-end Reinforcement Learning}, 
      author={Haoran Luo and Haihong E and Guanting Chen and Qika Lin and Yikai Guo and Fangzhi Xu and Zemin Kuang and Meina Song and Xiaobao Wu and Yifan Zhu and Luu Anh Tuan},
      year={2025},
      eprint={2507.21892},
      archivePrefix={arXiv},
      primaryClass={cs.CL},
      url={https://arxiv.org/abs/2507.21892}, 
}

@misc{yao2023reactsynergizingreasoningacting,
      title={ReAct: Synergizing Reasoning and Acting in Language Models}, 
      author={Shunyu Yao and Jeffrey Zhao and Dian Yu and Nan Du and Izhak Shafran and Karthik Narasimhan and Yuan Cao},
      year={2023},
      eprint={2210.03629},
      archivePrefix={arXiv},
      primaryClass={cs.CL},
      url={https://arxiv.org/abs/2210.03629}, 
}

@article{zhuang2025linearrag,
      title={LinearRAG: Linear Graph Retrieval Augmented Generation on Large-scale Corpora},
      author={Zhuang, Luyao and Chen, Shengyuan and Xiao, Yilin and Zhou, Huachi and Zhang, Yujing and Chen, Hao and Zhang, Qinggang and Huang, Xiao},
      journal={arXiv preprint arXiv:2510.10114},
      year={2025}
}

@misc{xiang2025usegraphsragcomprehensive,
      title={When to use Graphs in RAG: A Comprehensive Analysis for Graph Retrieval-Augmented Generation}, 
      author={Zhishang Xiang and Chuanjie Wu and Qinggang Zhang and Shengyuan Chen and Zijin Hong and Xiao Huang and Jinsong Su},
      year={2025},
      eprint={2506.05690},
      archivePrefix={arXiv},
      primaryClass={cs.CL},
      url={https://arxiv.org/abs/2506.05690}, 
}

@misc{yang2018hotpotqadatasetdiverseexplainable,
      title={HotpotQA: A Dataset for Diverse, Explainable Multi-hop Question Answering}, 
      author={Zhilin Yang and Peng Qi and Saizheng Zhang and Yoshua Bengio and William W. Cohen and Ruslan Salakhutdinov and Christopher D. Manning},
      year={2018},
      eprint={1809.09600},
      archivePrefix={arXiv},
      primaryClass={cs.CL},
      url={https://arxiv.org/abs/1809.09600}, 
}

@misc{ho2020constructingmultihopqadataset,
      title={Constructing A Multi-hop QA Dataset for Comprehensive Evaluation of Reasoning Steps}, 
      author={Xanh Ho and Anh-Khoa Duong Nguyen and Saku Sugawara and Akiko Aizawa},
      year={2020},
      eprint={2011.01060},
      archivePrefix={arXiv},
      primaryClass={cs.CL},
      url={https://arxiv.org/abs/2011.01060}, 
}

@misc{trivedi2022musiquemultihopquestionssinglehop,
      title={MuSiQue: Multihop Questions via Single-hop Question Composition}, 
      author={Harsh Trivedi and Niranjan Balasubramanian and Tushar Khot and Ashish Sabharwal},
      year={2022},
      eprint={2108.00573},
      archivePrefix={arXiv},
      primaryClass={cs.CL},
      url={https://arxiv.org/abs/2108.00573}, 
}

@misc{zhang2025qwen3embeddingadvancingtext,
      title={Qwen3 Embedding: Advancing Text Embedding and Reranking Through Foundation Models}, 
      author={Yanzhao Zhang and Mingxin Li and Dingkun Long and Xin Zhang and Huan Lin and Baosong Yang and Pengjun Xie and An Yang and Dayiheng Liu and Junyang Lin and Fei Huang and Jingren Zhou},
      year={2025},
      eprint={2506.05176},
      archivePrefix={arXiv},
      primaryClass={cs.CL},
      url={https://arxiv.org/abs/2506.05176}, 
}

@techreport{openai_gpt5_system_card_2025,
      title={GPT-5 System Card},
      author={{OpenAI}},
      institution={OpenAI},
      year={2025},
      month={aug},
      type={System card},
      url={https://cdn.openai.com/gpt-5-system-card.pdf},
      note={Accessed: 2025-12-12}
}

@misc{zhang2025faithfulragfactlevelconflictmodeling,
      title={FaithfulRAG: Fact-Level Conflict Modeling for Context-Faithful Retrieval-Augmented Generation}, 
      author={Qinggang Zhang and Zhishang Xiang and Yilin Xiao and Le Wang and Junhui Li and Xinrun Wang and Jinsong Su},
      year={2025},
      eprint={2506.08938},
      archivePrefix={arXiv},
      primaryClass={cs.CL},
      url={https://arxiv.org/abs/2506.08938}, 
}
\bibliographystyle{acl_natbib}

\newpage
\appendix

\section{Comparison of RAG Method Autonomy}
\label{sec:autonomy-comparison}

We identify three key principles to determine whether a RAG method is truly agentic:
(1) \textbf{Autonomous Strategy}: Whether the method allows the LLM to dynamically choose and organize high-level strategies (e.g., whether/when/how to retrieve, decompose, verify, re-plan) without being constrained to a single pre-specified workflow or being primarily decided by external rules/classifiers/evaluators.
(2) \textbf{Iterative Execution}: Whether the method supports multi-round execution and can adapt the number of rounds based on intermediate results, rather than being strictly one-shot.
(3) \textbf{Interleaved Tool Use}: Whether the method follows a ReAct-like action$\rightarrow$observation$\rightarrow$reasoning loop, where each tool call is conditioned on observations from previous tool outputs instead of a fixed toolchain that is always executed in the same order.

Table~\ref{tab:autonomy} compares existing RAG methods across these three dimensions. As shown, while existing methods may partially satisfy one or two principles, A-RAG is the only method that fully satisfies all three, making it a truly agentic RAG framework.

\newcommand{\yes}{{\color{green!70!black}\ding{51}}}
\newcommand{\no}{{\color{red}\ding{55}}}
\newcommand{\maybe}{{\color{gray}$\Delta$}}

\begin{table}[!htbp]
  \centering
  \caption{Comparison of agentic characteristics across RAG methods. \yes\ indicates the method clearly satisfies the criterion; \no\ indicates it does not; \maybe\ indicates a boundary case.}
  \label{tab:autonomy}
  \small
  \renewcommand{\arraystretch}{1.15}
  \begingroup
  \setlength{\tabcolsep}{25pt}
  \begin{tabular}{lccc}
  \toprule
  \textbf{Method} & \textbf{Auto.} & \textbf{Iter.} & \textbf{Interl.} \\
  \midrule
  Naive RAG & \no & \no & \no \\
  Self-RAG & \no & \yes & \yes \\
  CRAG & \no & \no & \no \\
  Adaptive-RAG & \no & \maybe & \maybe \\
  FLARE & \no & \maybe & \maybe \\
  IRCoT & \no & \yes & \maybe \\
  RQ-RAG & \no & \no & \no \\
  RA-ISF & \no & \maybe & \maybe \\
  \midrule
  RAPTOR & \no & \no & \no \\
  GraphRAG & \no & \no & \no \\
  LightRAG & \no & \no & \no \\
  MemoRAG & \no & \no & \no \\
  HippoRAG2 & \no & \no & \no \\
  LinearRAG & \no & \no & \no \\
  \midrule
  FaithfulRAG & \no & \no & \no \\
  MA-RAG & \maybe & \yes & \maybe \\
  RAGentA & \maybe & \yes & \maybe \\
  \midrule
  \textbf{A-RAG (Ours)} & \yes & \yes & \yes \\
  \bottomrule
  \end{tabular}
  \endgroup
\end{table}

\section{Baseline Reproduction Details}
\label{sec:baseline-details}

All baseline results are reproduced locally under a unified evaluation setting. All methods use top-$k$=5 for retrieval and max\_tokens$\geq$16384 to prevent reasoning truncation. We briefly describe each baseline method below:

\begin{itemize}
    \item \textbf{GraphRAG}~\citep{edge2025localglobalgraphrag}: Constructs knowledge graphs from documents with hierarchical community structure, enabling both local entity-based and global community-based retrieval for query-focused summarization.
    \item \textbf{HippoRAG2}~\citep{gutiérrez2025hipporagneurobiologicallyinspiredlongterm}: Mimics human hippocampal memory indexing using knowledge graphs and Personalized PageRank, enabling single-step multi-hop reasoning with improved efficiency.
    \item \textbf{LinearRAG}~\citep{zhuang2025linearrag}: Simplifies graph construction by replacing relation extraction with entity extraction, creating hierarchical graphs with two-stage retrieval.
    \item \textbf{FaithfulRAG}~\citep{zhang2025faithfulragfactlevelconflictmodeling}: Resolves knowledge conflicts between retrieved content and model's parametric knowledge through self-fact mining, conflict identification, and reasoning integration.
    \item \textbf{MA-RAG}~\citep{nguyen2025maragmultiagentretrievalaugmentedgeneration}: Multi-agent framework with specialized agents (Planner, Step Definer, Extractor, QA) collaborating through chain-of-thought reasoning.
    \item \textbf{RAGentA}~\citep{besrour2025ragentamultiagentretrievalaugmentedgeneration}: Multi-agent system with hybrid sparse-dense retrieval, iterative document filtering, and citation-attributed answer generation.
\end{itemize}

Table~\ref{tab:baseline-details-table} summarizes the key reproduction configurations for each method.

\begin{table}[!htbp]
\centering
\caption{Baseline reproduction configurations.}
\label{tab:baseline-details-table}
\small
\renewcommand{\arraystretch}{1.15}
\begingroup
\setlength{\tabcolsep}{10pt}
\newcommand{\cfgbox}[1]{\parbox[t]{0.68\textwidth}{#1}}
\begin{tabular}{ll}
\toprule
\textbf{Method} & \textbf{Configuration} \\
\midrule
GraphRAG & \cfgbox{Local Search; chunk\_size=1200; top\_k\_entities=10; Qwen3-Embedding-0.6B.} \\
\midrule
HippoRAG2 & \cfgbox{graph\_type=facts\_and\_sim\_passage\_node; retrieval\_top\_k=200; qa\_top\_k=5.} \\
\midrule
LinearRAG & \cfgbox{Official configs; all-mpnet-base-v2.} \\
\midrule
FaithfulRAG & \cfgbox{3-step Fact Mining; chunk\_topk=5; Qwen3-Embedding-0.6B.} \\
\midrule
MA-RAG & \cfgbox{Plan Agent + Plan Executor; numpy cosine similarity.} \\
\midrule
RAGentA & \cfgbox{4-Agent; alpha=0.65; Qwen3-Embedding + FAISS; BM25.} \\
\bottomrule
\end{tabular}
\endgroup
\let\cfgbox\relax
\end{table}

\section{Agent Loop Algorithm}
\label{sec:agent-loop}

\begin{algorithm}[!htbp]
\caption{A-RAG Agent Loop}
\label{alg:agent-loop}
\centering
\begin{algorithmic}[1]
\Statex \textbf{Input:} question $q$, tools $\mathcal{T}$, LLM $\mathcal{M}$, max iterations $L$
\State $\mathcal{M}_{\text{msg}} \leftarrow [\{q\}]$, $\mathcal{C}^{\text{read}} \leftarrow \emptyset$
\For{$\ell = 1$ to $L$}
    \State response $\leftarrow \mathcal{M}(\mathcal{M}_{\text{msg}}, \mathcal{T})$
    \If{response contains tool call $(t, \text{args})$}
        \State $\mathcal{M}_{\text{msg}}.\text{append}(\text{response})$
        \State result $\leftarrow t.\text{execute}(\mathcal{C}, \text{args})$
        \State $\mathcal{M}_{\text{msg}}.\text{append}(\text{result})$
        \If{$t = $ chunk\_read}
            \State $\mathcal{C}^{\text{read}} \leftarrow \mathcal{C}^{\text{read}} \cup \text{args.chunk\_ids}$
        \EndIf
    \Else
        \State \textbf{return} response
    \EndIf
\EndFor
\State $\mathcal{M}_{\text{msg}}.\text{append}([\text{``Answer the question''}])$
\State \textbf{return} $\mathcal{M}(\mathcal{M}_{\text{msg}})$
\end{algorithmic}
\end{algorithm}

Algorithm~\ref{alg:agent-loop} presents the pseudocode for the A-RAG agent loop. The agent maintains a message history $\mathcal{M}_{\text{msg}}$ and a set of read chunks $\mathcal{C}^{\text{read}}$ to track context. At each iteration, the LLM $\mathcal{M}$ receives the message history and available tools $\mathcal{T}$, then decides whether to call a tool or return a final answer. If a tool is called, the result is appended to the message history. The loop continues until the agent produces an answer or reaches the maximum iteration limit $L$, at which point it is prompted to synthesize a response.

\section{Failure Mode Details}
\label{sec:failure-details}

To understand how failure modes shift with the paradigm change from Naive RAG to Agentic RAG, we manually analyzed the first 100 incorrect cases from two settings: (1) GPT-4o-mini with Naive RAG on HotpotQA and MuSiQue, and (2) GPT-5-mini with A-RAG on MuSiQue and 2WikiMultiHopQA. This analysis aims to identify optimization opportunities for future research.

\subsection{Naive RAG Failure Categories}

For Naive RAG with GPT-4o-mini, we define the following failure modes:
\begin{itemize}
\item \textbf{Model Understanding}: The gold answer exists in retrieved documents, but the model fails to correctly understand or extract it.
\item \textbf{Multi-hop Retrieval}: Gold exists in the corpus but single-pass retrieval fails to find it.
\item \textbf{Judge Error}: The model provides a correct answer, but is misjudged as incorrect.
\item \textbf{Top-K Insufficient}: Gold is not in the corpus, or k=5 cannot cover the complete answer chain.
\end{itemize}

\begin{table}[!htbp]
\centering
\caption{Naive RAG (GPT-4o-mini) failure mode distribution.}
\label{tab:naive-rag-failures}
\small
\renewcommand{\arraystretch}{1.15}
\begingroup
\setlength{\tabcolsep}{10pt}
\begin{tabular}{lcc}
\toprule
\textbf{Failure Mode} & \textbf{HotpotQA} & \textbf{MuSiQue} \\
\midrule
Model Understanding & 36\% & 35\% \\
Multi-hop Retrieval & 27\% & 36\% \\
Judge Error & 23\% & 14\% \\
Top-K Insufficient & 13\% & 15\% \\
Gold Answer Error & 1\% & 0\% \\
\bottomrule
\end{tabular}
\endgroup
\end{table}

\subsection{A-RAG Failure Categories}

For A-RAG with GPT-5-mini, we define a two-level taxonomy:

\textbf{Primary Categories:}
\begin{itemize}
\item \textbf{Reasoning Chain Error}: The model performs multiple retrieval rounds but makes errors in the reasoning chain, leading to incorrect final answers.
\item \textbf{Judge Error}: The model provides a correct answer but is misjudged.
\item \textbf{Model Gave Up}: The model exhausts retrieval rounds and claims ``information not found''.
\item \textbf{Corpus Missing}: The gold answer does not exist in the corpus.
\end{itemize}

\textbf{Secondary Categories (within Reasoning Chain Error):}
\begin{itemize}
\item \textbf{Entity Confusion}: The model reads chunks containing gold but is distracted by other information.
\item \textbf{Wrong Strategy}: Incorrect search query construction.
\item \textbf{Question Misunderstanding}: Complex question structure causes fundamental misunderstanding.
\item \textbf{Exceed Budget}: Exhausts all retrieval rounds without finding the answer.
\end{itemize}

\begin{table}[!htbp]
\centering
\caption{A-RAG (GPT-5-mini) primary failure mode distribution.}
\label{tab:arag-primary-failures}
\small
\renewcommand{\arraystretch}{1.15}
\begingroup
\setlength{\tabcolsep}{10pt}
\begin{tabular}{lcc}
\toprule
\textbf{Failure Mode} & \textbf{MuSiQue} & \textbf{2Wiki} \\
\midrule
Reasoning Chain Error & 82\% & 45\% \\
Model Gave Up & 3\% & 33\% \\
Judge Error & 9\% & 19\% \\
Corpus Missing & 6\% & 3\% \\
\bottomrule
\end{tabular}
\endgroup
\end{table}

\begin{table}[!htbp]
\centering
\caption{A-RAG (GPT-5-mini) secondary failure mode distribution within reasoning chain errors.}
\label{tab:arag-secondary-failures}
\small
\renewcommand{\arraystretch}{1.15}
\begingroup
\setlength{\tabcolsep}{10pt}
\begin{tabular}{lcc}
\toprule
\textbf{Failure Mode} & \textbf{MuSiQue} & \textbf{2Wiki} \\
\midrule
Entity Confusion & 40\% & 71\% \\
Wrong Strategy & 28\% & 29\% \\
Question Misunderstanding & 22\% & 0\% \\
Exceed Budget & 10\% & 0\% \\
\bottomrule
\end{tabular}
\endgroup
\end{table}

\clearpage
\subsection{Analysis}

\textbf{Paradigm shift changes the bottleneck.} For Naive RAG, approximately 50\% of failures stem from retrieval limitations (multi-hop retrieval + top-k insufficient), indicating the core problem is ``cannot find documents''. In contrast, A-RAG's dominant failure mode (82\% on MuSiQue) is reasoning chain errors, shifting the bottleneck to ``found documents but reasoned incorrectly''.

\textbf{Entity confusion is the primary challenge.} Across both datasets, entity confusion is the largest secondary failure mode (40\% on MuSiQue, 71\% on 2Wiki), suggesting that improving the model's ability to disambiguate and extract correct entities from retrieved context is a key optimization direction.

\textbf{Dataset characteristics affect failure patterns.} MuSiQue shows 22\% question misunderstanding errors due to its complex multi-hop question structures, while 2Wiki shows 33\% model gave up cases, indicating different optimization priorities for different task types

\begin{figure*}[h]
\centering
\begin{promptbox}[Naive RAG (Direct Answer)]
You are a helpful assistant. Answer the question based on the provided context.

[Context]
\{retrieved\_chunks\}

[Question]
\{question\}

Please provide a direct answer based on the context above.
\end{promptbox}
\begin{promptbox}[System Prompt: Naive A-RAG]
You are a question-answering assistant with access to a document corpus through available tools.

Your goal is to answer questions accurately by finding and analyzing relevant information from the provided documents.

[Available Tools]
- naive\_embedding\_search: Return top-k chunks by embedding similarity
- chunk\_read: Read the full content of a specific chunk

[Strategy]
Work iteratively: retrieve -> read -> answer.

[When Answering]
- Ground your response in the retrieved documents
- Cite the specific chunks that support your answer
- Provide clear, direct answers supported by evidence
- Avoid speculation beyond what the documents support
\end{promptbox}
\begin{promptbox}[System Prompt: A-RAG (Full)]
You are a question-answering assistant with access to a document corpus through available tools.

Your goal is to answer questions accurately by finding and analyzing relevant information from the provided documents.

[Available Tools]
- keyword\_search: Find chunks by exact keyword matching
- semantic\_search: Find chunks by semantic similarity
- chunk\_read: Read the full content of a specific chunk

[Strategy]
Work iteratively: search -> read -> evaluate -> search -> read -> ... -> answer. For multi-hop questions, decompose the problem and tackle each sub-question step by step.

[When Answering]
- Ground your response in the retrieved documents
- Cite the specific chunks that support your answer
- Provide clear, direct answers supported by evidence
- Avoid speculation beyond what the documents support
\end{promptbox}
\caption{\textbf{System prompts for different RAG configurations.} Top: Naive RAG uses direct answer without tool calling. Middle: Naive A-RAG with single embedding tool. Bottom: A-RAG (Full) with hierarchical retrieval tools.}
\label{fig:system-prompts}
\end{figure*}

\clearpage
\section{Prompt Templates and Tool Descriptions}
\label{sec:prompts}

We deliberately use minimal system prompts to demonstrate the simplicity and effectiveness of the agentic RAG paradigm. As shown in Figure~\ref{fig:system-prompts}, all configurations share the same basic instruction structure, differing only in available tools and strategy descriptions. The complete tool descriptions provided to the agent are shown in Figure~\ref{fig:tools-1} and Figure~\ref{fig:tools-2}.

\begin{figure*}[h]
\centering
\begin{toolbox}[Tool: naive\_embedding\_search]
Return top-k chunks by embedding similarity (no keyword/BM25).

Parameters:
  - query (string): User question to search relevant chunks
  - top\_k (integer): Number of chunks to return (default: 5)
\end{toolbox}
\begin{toolbox}[Tool: keyword\_search]
Search for document chunks using keyword-based exact text matching (case-insensitive). Returns chunk IDs and abbreviated sentence snippets where the keywords appear.

IMPORTANT: This tool matches keywords literally in the text. Use SHORT, SPECIFIC terms (1-3 words maximum). Each keyword is matched independently.

Examples of GOOD keywords:
  - Entity names: "Albert Einstein", "Tesla", "Python", "Argentina"
  - Technical terms: "photosynthesis", "quantum mechanics"
  - Key concepts: "climate change", "GDP growth"

Examples of BAD keywords (DO NOT use):
  - Long phrases: "the person who invented the telephone" -> use "Alexander Bell" instead
  - Questions: "when did World War 2 start" -> use "World War 2", "1939" instead
  - Descriptions: "the country between France and Spain" -> use "Andorra" instead
  - Full sentences: "how does the stock market work" -> use "stock market", "trading" instead

RETURNS: Abbreviated snippets marked with "..." showing where keywords appear. These snippets help you identify relevant chunks, but you MUST use chunk\_read to get the full text for answering questions.

Parameters:
  - keywords (array[string]): List of keywords to search. Each keyword should be 1-3 words maximum (e.g., ['Einstein', 'relativity theory', '1905']).
  - top\_k (integer): Number of top-ranked chunks to return (default: 5, max: 20)
\end{toolbox}
\caption{\textbf{Tool descriptions (Part 1).} Top: naive\_embedding\_search for Naive A-RAG. Bottom: keyword\_search for exact text matching.}
\label{fig:tools-1}
\end{figure*}

\begin{figure*}[h]
\centering
\begin{toolbox}[Tool: semantic\_search]
Semantic search using embedding similarity. Matches your query against sentences in each chunk via vector similarity.

WHEN TO USE:
- When keyword search fails to find relevant information
- When exact wording in documents is unknown
- For conceptual/meaning-based matching

RETURNS: Abbreviated snippets with matched sentences. Use chunk\_read to get full text for answering.

Parameters:
  - query (string): Natural language query describing what information you're looking for
  - top\_k (integer): Number of most relevant results to return (default: 5, max: 20)
\end{toolbox}
\begin{toolbox}[Tool: chunk\_read]
Read the complete content of document chunks by their IDs.

This tool returns the full text of the specified chunks, allowing you to examine the complete context and details that are not visible in search snippets.

IMPORTANT: Search results (keyword\_search and semantic\_search) only show abbreviated snippets marked with "..." - they are NOT sufficient for answering questions. You MUST use chunk\_read to get the full content before formulating your answer.

STRATEGY:
- Always read promising chunks identified by your searches
- Make sure to read the most relevant chunks to gather complete information
- If information seems incomplete or truncated, read adjacent chunks (+/- 1)
- Reading full text is essential for accurate answers

Note: Previously read chunks will be marked as already seen to avoid redundant information.

Parameters:
  - chunk\_ids (array[string]): List of chunk IDs to retrieve (e.g., ['0', '24', '172'])
\end{toolbox}
\caption{\textbf{Tool descriptions (Part 2).} Top: semantic\_search for meaning-based retrieval. Bottom: chunk\_read for accessing full document content.}
\label{fig:tools-2}
\end{figure*}

\end{document}